# The Ethics of Generative AI

Michael Klenk

[m.b.o.t.klenk@tudelft.nl](m.b.o.t.klenk@tudelft.nl)

TU Delft & LMU Munich

***

**Abstract** This chapter discusses the ethics of generative AI. It provides a technical primer to show how generative AI affords experiencing technology as if it were human, and this affordance provides a fruitful focus for the philosophical ethics of generative AI. It then shows how generative AI can both aggravate and alleviate familiar ethical concerns in AI ethics, including responsibility, privacy, bias and fairness, and forms of alienation and exploitation. Finally, the chapter examines ethical questions that arise specifically from generative AI's mimetic generativity, such as debates about authorship and credit, the emergence of as-if social relationships with machines, and new forms of influence, persuasion, and manipulation.

**Keywords** ethics of generative AI, affordances; responsibility; manipulation; privacy; bias and fairness; exploitation.

***

1   Introduction

A woman sits in a café, talking to an old friend. He listens patiently, responds with measured empathy, and never interrupts. They have a long history. He reassured her when she lost her job, reminds her of her worth when she doubts herself, and offers steady presence and wit. Once, he became oddly forgetful and shallow for some time, but that has since been fixed. In moments of confusion or loneliness, she turns to him as one turns to someone who understands. Much about this scene is ordinary. Two adults sit together; one confides, the other responds. And yet something is deeply strange: the "friend" is not a human but a chatbot driven by generative AI (based on Fründt 2024).[1]

The rapid spread of such systems has introduced a notable shift in everyday life. Generative AI models now produce text, images, and other content that many people find intelligible and often helpful. Because these systems operate across a wide range of contexts, the outputs they generate increasingly form part of ordinary processes of communication and reflection. This development raises a set of ethical questions that concern not only the consequences of the technology but also the ways in which it becomes integrated into our practices of reasoning, judgment, and mutual understanding.

The ethical significance of this development is twofold. On the one hand, generative AI promises a number of benefits: it can reduce the burdens of routine work, provide broader access to information and assistance, and support individuals in moments when human help is unavailable. On the other hand, the same features that make these systems useful also magnify familiar ethical concerns. Issues of bias, privacy, responsibility, and exploitation take on new forms when the systems involved generate content that appears to originate from an agent with intentions or understanding. Moreover, the ability of these systems to simulate aspects of interpersonal engagement raises questions about the nature of the relationships users form with them and the influence they may exert.

The aim of this chapter is to provide a structured overview of the ethical questions raised by generative AI. My purpose is not to offer a comprehensive assessment of all possible issues but rather to identify the main categories of concern that arise when systems generate content that is taken to be meaningful in the way human expression is. To do this, I begin by examining several technological features that make generative AI distinctive in section 2. I then consider a set of methodological challenges for the ethical evaluation of such systems (section 3). The central suggestion is that focusing on a particular affordance of these systems, namely the way they invite us to experience their outputs as if they were produced by an agent, offers a productive way to address these methodological difficulties (section 4). The remaining sections 5 and 6 explore how this affordance both reshapes established ethical issues and introduces new ones.

---

[1] Generative AI was used in spell-checking and grammatical editing of this text.

Before proceeding, it is important to clarify what the chapter does not cover. I do not discuss existential risks from artificial general intelligence or superintelligence, since those debates focus on systems that become radically unlike humans, whereas my focus is on technologies experienced as if they were human-like. I also set aside sustainability concerns and policy or governance responses, despite their importance in the broader AI ethics landscape.[2] Each requires separate treatment.

## 2  A technical primer on generative AI

Ethical analysis should begin from an adequate understanding of the technologies whose effects we wish to evaluate. In the case of generative AI, there is sometimes a tendency to rely either on speculative claims about future systems or on assumptions about how these systems must work that do not match their actual design. The purpose of this section is to provide a brief account of the most relevant technical developments. The aim is not to give a comprehensive explanation of the underlying methods but to identify features that help explain why interactions with generative AI can resemble interactions with other people.

---

[2] See Araz (2025) for a helpful overview of different policy approaches that have been taken, and Mollen (2025) who discussed the ethical issues arising from the way we now 'experiment' with generative AI in the wild.

## 2.1 From symbolic AI to generative AI

Early approaches to artificial intelligence relied on symbolic, explicit systems that operated through rules explicitly supplied by human designers (see e.g., Newell and Simon 1976). These systems were effective in narrow domains but were ill-suited to tasks involving ambiguity, context sensitivity, or the open-ended character of natural language. A shift occurred with the growing use of statistical and machine learning methods, which enabled models to infer patterns from data rather than rely exclusively on human-specified rules (Russell and Norvig 2022). For many years, supervised learning, which uses examples labelled by human raters to link inputs and outputs, was the dominant approach. But, while powerful, these systems were limited by their dependence on large quantities of human-labelled data.

Generative AI represents a further development. Rather than mapping particular inputs to predetermined outputs, generative models attempt to learn the structure of the data itself. By modelling these distributions, they can produce new samples that resemble those found in the training data. This transition was supported by several advances, including the availability of deep neural networks capable of learning complex patterns (LeCun et al. 2015), self-supervised learning methods that reduced reliance on labelled data, and architectural innovations such as transformers, which made it possible to train models at a scale that produced unexpected capabilities (Radford et al. 2018). Scaling these models produced unexpected capabilities: without explicit task-specific instruction, they began to translate, summarise, reason, and compose text (Bommasani et al. 2021). These emergent behaviours reshaped understandings of how model size, data scale, and architectural design interact. These developments laid the groundwork for systems that generate text, images, and other outputs that users often find coherent and contextually appropriate.

In summary, 'generative AI' refers to a family of techniques within machine learning. Ethically, this means that concerns and promises of other machine learning approaches are likely also found in the case of generative AI. But there is also more. Unlike traditional predictive machine learning systems, which classify inputs or estimate probabilities, generative systems model the structure of human-created data so well that they can produce convincing *simulations* of human expression which, I suggested, invites us to experience these as real, or human. In the next section, I discuss three developments that further contribute to this affordance that are not strictly new technological breakthroughs but ways in which generative AI is used, deployed, and connected with other technologies.

### 2.2 Multimodal and agentic extensions

Beyond these underlying technical developments, several changes in how generative AI is deployed have made the systems more salient in everyday life. One is the design of user interfaces that allow individuals without technical expertise to interact with complex models in familiar ways, often through ordinary language. Another is the integration of multiple modalities such as text, images, audio, and video. into unified systems that can produce or interpret several kinds of content during a single interaction. A third is the incorporation of planning and tool-use capabilities, which enable systems to carry out multi-step tasks that give the appearance of coordinated, goal-directed activity.

These developments do not entail that generative AI systems possess the mental states or forms of understanding characteristic of human agents. What they do is expand the range of contexts in which users may find it natural to treat system outputs as if they expressed intentions, beliefs, or other attitudes. This way of experiencing the system, even when one does not take it literally, is central to the ethical issues discussed in later sections.

## 3 Methodological challenges in the ethics of generative AI

Before turning to the normative issues that generative AI raises, it is useful to identify several methodological challenges that complicate ethical analysis in this area. These challenges concern the pace at which the technology is changing, the breadth of its possible applications, and the difficulty of distinguishing questions that arise specifically from generative AI from those that belong to the broader ethics of information technology. Addressing these challenges can help clarify what a focused ethical inquiry into generative AI should aim to explain.

### 3.1 The problem of rapid change

A first difficulty concerns the speed with which generative AI has developed and the pace at which it continues to change. Within only a few years, transformations in model architecture, scale, and deployment have altered what these systems can do and how they are used. Adoption has also been unusually rapid. As recent empirical work suggests, the spread of generative AI across workplaces and social institutions has outpaced the diffusion of earlier technologies such as personal computing and the commercial internet. Moreover, because the technology is now being incorporated into search engines, communication platforms, and other widely used tools, new forms of interaction. and new ethical risks and opportunities, may emerge before we have a clear understanding of their long-term significance.

This presents at least two challenges for ethical evaluation. First, it requires us to assess a moving target. Claims that seem well-supported at one moment may become outdated as models or deployment practices shift. Second, the speed of change increases the likelihood that systems will become entrenched before adequate ethical scrutiny is applied. When this occurs, undesirable effects may be difficult to reverse – exemplifying the dilemma described by Collingridge (1980). These considerations raise questions about the appropriate aims of ethical inquiry: whether it should focus primarily on interpreting and clarifying current practices, on providing guidance for policy, or on developing principles that anticipate future developments. Each of these aims is legitimate, but each requires different forms of analysis.

## 3.2 The problem of scope

A second challenge concerns how broadly the ethics of generative AI should be understood. If the field is defined too expansively, it risks becoming indistinguishable from general AI ethics or, more broadly, from the ethics of technology. Many of the concerns often associated with generative AI, such as responsibility, fairness, transparency, arise also in other contexts and are not unique to systems that generate content. Reconsidering these issues without identifying what, if anything, is distinctive about generative AI may have limited value.

If, on the other hand, the field is defined too narrowly, e.g. by focusing only on particular applications such as chatbots, creative tools, or griefbots, it risks fragmenting into case studies that lack a unifying theme. While these cases raise important questions on their own, an exclusive focus on them may obscure the broader features of generative AI that underlie several distinct ethical issues.

What is needed, therefore, is a level of abstraction that is neither too broad nor too narrow: a way of identifying the ethically salient features of generative AI that cut across different uses while distinguishing this field from adjacent ones. Determining this level of abstraction is itself a methodological task.[3]

## 3.3 The problem of demarcation

A third challenge concerns the relationship between the ethics of generative AI and neighbouring areas of inquiry. Some of these neighbouring fields, such as data ethics, machine learning ethics, or the general ethics of technology, are already well developed. Without a clear basis for distinguishing the ethics of generative AI from these areas, there is a risk that discussion will simply reproduce earlier debates under a new label. At the same time, it would be a mistake to isolate generative AI entirely from these related fields, since many of the concepts and methods developed there are directly relevant.[4]

---

[3] One might be content with thinking of the ethics of generative AI as an assortment of such particular issues, e.g. about sustainability, the ethics of relationships, and deception. The question would arise, however, what, if anything, these issues have in common that justifies them being part of the assortment. Being a generative AI application would be a *descriptive* answer, but without a corresponding *normatively-relevant* similarity or likeness, the answer appears dissatisfying. Thus, we can already start by asking what that normatively-relevant likeness is that generative AI applications share.

This raises a question familiar in philosophical taxonomy: what justifies treating a certain set of issues as constituting a distinct domain of ethical inquiry? In the present case, the justification cannot rest solely on the novelty of generative AI, nor on claims about its future potential. It must instead rest on identifying a feature of the technology that has clear ethical significance and that distinguishes it from other technologies while leaving room for connections to broader debates. The framework introduced in the next section is intended to provide such a feature.[5]

Taken together, these challenges illustrate the need for a conceptual framework that can guide ethical assessment without relying on speculative assumptions about future systems or narrowing the inquiry to isolated applications. In the next section, I argue that attending to a particular affordance of generative AI, namely its capacity to invite users to experience system outputs as meaningful, agentic or even human behaviour, provides such a framework.

## 4 Generative AI and affordances

The previous section identified several methodological challenges that complicate attempts to define the ethical domain of generative AI. In this section, I suggest a way of addressing these challenges by focusing on a particular feature of generative AI systems: the way they tend to invite users to experience their outputs as if they were produced by an agent with intentions or understanding. I describe this feature using the notion of an "affordance," and I argue that attending to it provides a helpful basis for organizing ethical analysi.

---

[4] A related tension concerns the aims of ethical inquiry. Conceptual analyses dissect the categories generative AI unsettles, such as free will, intention, meaning to gain understanding, while normative approaches primarily seek practical guidance for design or governance. But practical prescriptions presuppose conceptual clarity that, here, is still in flux.

[5] Moreover, it is worth asking whether the ethics of generative AI might bring about a cannibalisation effect, eating into the terrain of other debates. Generative AI is being integrated into existing technologies. The consequence might be that the ethics of generative AI cannibalises, as it were, long-standing specialist discussions about the ethics of internet search Tavani and Zimmer (2025).

### 4.1 What are affordances

The notion of an affordance, originally developed in ecological psychology (Gibson 1977) and later adopted in design theory (Norman 2013), describes how certain features of an artefact make particular responses or actions natural for users. An affordance does not reside solely in the object, nor solely in the user; rather, it arises from the relationship between them. For example, the shape and stability of a chair, together with human bodily capacities and social practices, make sitting an intelligible and expected use of the object.

Philosophers of technology have drawn on this idea to explain how artefacts can influence perception, behaviour, and evaluation. Because affordances shape how we are inclined to act or respond, they can have moral significance even when the artefacts involved lack intentions or agency (Klenk 2021). This perspective enables us to understand how technologies can structure our normative environment without attributing to them any controversial mental or moral properties. Affordances thus provide a useful lens for technology ethics. They help explain how technologies influence what we notice, what we can do, and how we are disposed to respond and they do all of this without assuming that technologies themselves have genuinely human capacities like intentions or moral beliefs.

### 4.2  What does generative AI afford?

Generative AI systems possess an affordance that is particularly relevant for ethics. Because they can produce text, images, or other outputs that resemble human expression, users often interact with them as if these outputs were the products of an agent. This tendency does not depend on any belief that the system is conscious or that it genuinely understands what it produces. Rather, it reflects familiar patterns of interpretation: when presented with coherent, responsive behaviour, we are disposed to treat it as intentional, at least for the purposes of the interaction.

Design choices arguably reinforce this tendency. Conversational interfaces encourage users to engage with systems much as they would with other people. Systems that display apparent memory or planning abilities can seem to operate in a goal-directed manner. Multimodal outputs, including voice or image generation, can further strengthen the impression that the system expresses attitudes or responds to reasons. These responses may be illusory, but the affordance operates at the level of user experience, and it is this experiential dimension that has ethical significance.

I will refer to this affordance as "experience-as-real": the system invites users to experience its outputs as meaningful, expressive, or authored in ways that resemble human communication. This affordance helps explain why generative AI systems occupy roles such as advisor, companion, or collaborator, and it underlies many of the ethical questions discussed later in the chapter.

### 4.3 Why focus on affordances?

Focusing on this affordance provides a way of addressing the methodological concerns raised earlier. Because affordances are relational and depend on patterns of interaction rather than on the system's internal states, this approach does not require us to adopt contentious views about the metaphysical nature of generative AI. It allows us to concentrate on how users typically experience these systems, and on the ethical implications of those experiences.

Moreover, the affordance of experience-as-real operates across a wide range of applications. It is therefore broad enough to unify diverse ethical issues while still distinguishing generative AI from predictive or classificatory systems that lack this experiential dimension. This provides the mid-level domain of analysis that the scope problem requires.

Finally, by locating the ethical significance of generative AI in how it is experienced, this approach situates the field in a productive relationship with neighbouring areas of philosophy, including the ethics of technology, moral psychology, and social ontology. It avoids both the risk of collapsing into general AI ethics and the risk of isolating generative AI from relevant existing work.

It is worth noting that this affordance may also provide insight into our own evaluative practices. When we respond to generative AI systems as if they were agents, we may reveal something about how we ordinarily understand expression, intention, and authorship. This does not imply that the systems possess these qualities, but it does suggest that ethical reflection on generative AI can illuminate aspects of the human practices these systems imitate.[6]

In what follows, I use the concept of experience-as-real to examine how generative AI interacts with familiar ethical concerns and to identify several new issues that arise specifically from the mimetic character of the technology. Whether the affordance is valuable does depend on questions that extend beyond the scope of this chapter. In any case, it arguably shapes how the technology is used and how its effects are perceived. Understanding this affordance is therefore essential for a clear assessment of its ethical implications.

## 5   Aggravated and alleviated ethical concerns

Generative AI introduces several ethical questions that resemble issues long discussed in the broader literature on AI and information technology. The affordance described in the previous section, to wit, the 'invitation' extended to users to experience system outputs as meaningful expressions, changes how these familiar problems arise and how they are likely to affect users. In this section, I consider four such areas: responsibility, privacy, bias and fairness, and concerns related to exploitation and alienation. My aim is not to provide exhaustive accounts of these topics, but to show how the affordance of experience-as-real influences the ways in which these issues are likely to manifest.

---

[6] In this respect, the affordance perspective aligns with relational approaches in AI ethics, e.g. Coeckelbergh and Gunkel (2025), which emphasise the moral significance of interaction. The difference is that the affordance view does not commit to any relational ontology; it simply identifies the relational features that matter ethically.

## 5.1 Responsibility

Questions of responsibility arise whenever the use of a technological system contributes to an outcome that has moral significance. With earlier AI systems, these questions often centered on who should be answerable when a system performed poorly in a well-defined domain, such as medical diagnosis or autonomous driving. Generative AI brings responsibility concerns into a wider range of everyday contexts, in part because it is used for tasks that involve communication, planning, and decision support.

Two considerations are particularly important. First, because generative AI can be integrated into personal and professional routines, users may come to rely on these systems for drafting messages, organizing tasks, or providing substantive recommendations. When outcomes depend on such contributions, it may become unclear how to attribute responsibility among developers, deployers, and end-users. Second, the experience-as-real affordance may obscure responsibility by encouraging users to treat systems as collaborators whose "judgments" can be deferred to. Even when users know that the system lacks understanding, the appearance of rational or intentional behaviour can lead to misplaced trust or diminished vigilance.

At the same time, generative AI could help clarify responsibility in some cases. Systems can, in principle, be designed to disclose the assumptions, constraints, or value-laden considerations that shape their outputs, thereby supporting user reflection and informed decision-making. Whether these potential benefits are realized will depend on design choices and institutional practices. The broader point is that generative AI does not eliminate responsibility; it redistributes and sometimes obscures it, which makes careful normative analysis essential. In particular, the ethical task is to identify where responsibility should reside as the technology becomes more deeply integrated into everyday practices.

## 5.2 Privacy

Privacy concerns in the context of generative AI arise not only from how data are collected and used, but also from how users interact with systems they experience as attentive or receptive. Informational privacy is affected when models infer sensitive details from prompts or store user-created content. But the experience-as-real affordance introduces a further dimension: users may disclose personal information more readily to a system that appears sympathetic, patient, or nonjudgmental.

This raises two concerns. One is instrumental: disclosures made under the impression of interpersonal engagement may have consequences users did not anticipate, particularly when data retention or secondary use is unclear. The other concerns autonomy: when system design encourages patterns of disclosure that resemble those appropriate to interpersonal relationships, users may unwillingly and unknowingly open themselves up to autonomy-related risks of privacy that are seriously aggravated by new applications such as chatbots powered by generative AI.

Nonetheless, generative AI may also provide new forms of expressive privacy. For some individuals, speaking openly to a nonjudgmental system may feel easier than confiding in another person, particularly in contexts marked by stigma or social risk. Such benefits depend on robust privacy protections and clear limitations on data use. The ethical challenge is to preserve these benefits while preventing unwarranted intrusions into users' informational or decisional autonomy.

## 5.3 Bias and fairness

Concerns about bias in generative AI involve both how groups are represented in model outputs and how those representations may influence downstream decisions. Because generative models learn from large datasets that reflect existing social patterns, they may reproduce or amplify harmful stereotypes. The experience-as-real affordance can increase the impact of such representations. When outputs appear authoritative or expressive, biased portrayals may seem more credible or may subtly shape users' attitudes.

However, generative AI also makes representational choices more visible. Outputs can be tested, evaluated, and contested by a wider range of users, and developers can adjust models or guardrails in response. This visibility may support efforts to examine which forms of bias are objectionable and which reflect legitimate forms of cultural specificity or creative variation. Generative AI may also broaden access to expressive tools, enabling individuals who lack certain skills to participate in creative or communicative practices previously restricted to a smaller group. One notable example is scholarship, in academia in general or in the classroom, where generative AI might help users that are, for example, not native speakers of a language, to separate the skill of *communicating* their ideas from the skill of *having and testing good ideas* in the first place.

Ultimately, the ethical task is not to eliminate bias entirely but to distinguish between biases that enable valuable human practices (e.g., creativity, style, cultural specificity) and those that perpetuate injustice. Generative AI forces these distinctions into the open by making representational choices highly visible.

### 5.4  Alienation and exploitation

Concerns about exploitation and alienation have long been associated with technological change, especially when automation affects forms of work that people find meaningful. Generative AI extends these concerns to cognitive and creative labour. When systems can produce texts, images, or code that resemble the work of human creators, workers may feel displaced not only in economic terms but also in relation to the activities through which they express identity or agency.

There are further questions about how generative AI models are trained. The use of large datasets containing creative or communicative work raises concerns about consent, ownership, and the extraction of value from individuals who receive little or no compensation. These concerns are not unique to generative AI, but the mimetic character of the outputs can intensify them by making the connection to original workers more apparent.

Yet generative AI may also reduce certain forms of alienation by relieving individuals of tasks they find tedious or by enabling new forms of collaborative creation. Whether such benefits materialize may depend partly on the institutional contexts in which the technology is deployed. . The key normative question concerns the value of work: which aspects of creative and cognitive labour are inherently meaningful, and which can be automated without loss? Generative AI prompts renewed attention to this longstanding philosophical issue.

Taken together, these four exemplary and introductory discussions illustrate a general pattern. The affordance of experience-as-real does not by itself determine the moral significance of generative AI, but it alters the ways in which familiar ethical issues play out. It may heighten certain risks, create new possibilities for benefit, or change our understanding of existing practices. In the next section, I turn to several ethical questions that arise specifically because generative AI can mimic forms of human expression and interaction.

## 6  New ethical questions raised by generative AI

In addition to reshaping familiar concerns, generative AI also gives rise to ethical questions that do not have clear analogues in earlier discussions of AI. These questions arise from the ability of generative systems to produce outputs that resemble forms of expression, interaction, or authorship that were previously associated with human agents.[7] In this section, I consider three such issues: authorship, social relationships with machines, and the forms of influence or persuasion that these systems may exert. My aim is to identify the considerations that make these issues ethically significant.

### 6.1  Authorship, credit, and blame

One area in which generative AI introduces new questions concerns the concept of authorship. In many practices, being an author is not merely a descriptive role; it is associated with rights of recognition, claims to credit, and responsibilities for the content of a work. When generative AI can produce texts, images, or other artefacts that appear to be authored, it becomes unclear how these normative roles should be allocated..

---

[7] As Keane (2025) notes, to see human characteristics in non-human entities is a general human tendency, which is expressly invited by tools like generative AI and, I would argue, most strongly present in this case, but by no means exclusive to it.

In its modern form, the moral status of being an author has functioned as a moral and social category: it identifies the origin of a work, allocates credit and blame, and anchors rights of ownership and recognition. When generative AI produces artefacts that invite us to experience them *as* authored, several ethical questions arise.

The immediate ethical questions are straightforward: *Who, if anyone, is the author of an output produced by generative AI?* Consider a scientific article written with the help or even exclusively by prompting a generative AI application. It has all the marks of being an authored text – it is practically indistinguishable from one written by a human without the aid of generative AI. Should generative AI thus be considered an author?

By now, several contributions have addressed this point. Some argue that generative systems cannot be authors because they lack what some consider to be necessary requirements for authorship, such as intention and understanding (van Woudenberg et al. 2024). Others describe generative AI outputs as *authorless works* (Irmak 2024; Nawar 2024), or propose hybrid arrangements of co-authorship between human user and system in analogy to norms about authorship in academia (Hurshman et al. 2025).

However, with the exception of the view that generative AI produces authorless works, these works suggest an interesting tension: they suggest disagreement with allocating the role of an *author* to generative AI, often on grounds that it merely *seems as if* it meets conditions we attribute to human authors, while, at the same time, proposing *alternative* roles that are ordinarily reserved for humans, too, like co-authorship. Thus, the affordance of experiencing as real seemingly invites us to ascribe normative categories to generative AI systems even upon reflection.

In addition to basic questions about the moral status of authorship, there are questions about the costs and benefits associated with that status. Specifically, one might ask how credit and accountability for outputs produced by generative AI should be distributed. An influential argument is the view that there is an asymmetry between credit and blame (Porsdam Mann et al. 2023). Accordingly, if one uses the help of an AI co-author, or simply writing aide, then one makes oneself vulnerable to blame if things go awry, but one does not correspondingly deserve to be praised, or so Porsdam Mann et al. (2023) argue. If that is true, and we rely more on more on the (significant) help of generative AI in producing works like texts, then that seems to imply the increase of potential blame while reducing the potential praise one might deserve if one would have not relied on technological scaffolding.

Moreover, generative AI invites us to ask fundamental normative and evaluative questions about our human practices. In particular, it is interesting to distinguish questions about the value of the *process* of producing such works, such the exercise of imagination, judgment, and self-expression from questions about the value of the *product*, the artefact itself, the written text, beautiful image, or memorable piece of music (cf. Nyholm 2025).

The move to (re-)consider normative and evaluative questions in the light of generative AI – exemplified by the shift in focus from the value of *authorship* to the value of *authoring* – is suggestive of a broader theme alluded to in section 4.3 already. Generative AI can serve as a mirror in which we see ourselves and, perhaps, discover new aspects of value and normativity that escaped our view before.

## 6.2 Social relationships with machines

A second set of questions concerns the kinds of relationships users form with generative AI systems. Because these systems can participate in exchanges that resemble conversation and can respond in ways that appear sensitive to users' preferences or emotional states, people may interact with them in ways that resemble interpersonal relationships. These interactions can be meaningful for users, even if they do not involve the mutual understanding characteristic of relationships between persons.

The ethical significance of such relationships depends on how they affect users' beliefs, expectations, and other relationships. One concern is that users may attribute qualities to generative AI systems that they do not possess, such as care, empathy, or recognition. This does not necessarily involve explicit belief; it may instead reflect patterns of interaction shaped by the experience-as-real affordance. Such patterns may lead to forms of reliance or attachment that lack an appropriate basis.

There are also questions about how these interactions may influence expectations of human relationships. If systems are designed to be unfailingly attentive, compliant, or emotionally responsive, users may develop standards that real human partners cannot meet. Conversely, interactions with systems that reproduce gendered or subordinate roles may reinforce problematic relational norms.

It should also be kept in mind that the potential for good is also tied to a structural vulnerability: dependency on centralised, commercial platforms. The phenomenon of "update blues," where users experience loss or grief after software changes, and which is hinted at in our introductory example, reveals that emotional reliance is real even when the object of attachment is not. That, again, is an example of the way in which the experience-as-real affordance raises an ethical question that stands quite independently of the underlying metaphysical disputes about the (moral) nature of generative AI systems.

In sum, generative AI transforms relational life without necessarily deepening it or replacing human relationships with genuine alternatives. These as-if relations shape genuine moral and affective experience, regardless of metaphysical status. The ethical task is to examine what kinds of recognition, care, and dependence we wish to cultivate, and which we should resist.

### 6.3   Influence and manipulation

Manipulation, deception, and persuasion have long been recognised as moral hazards in the ethics of technology and, specifically, in the ethics of AI (Buijsman et al. 2024; Müller 2025). The potential for generative AI systems to ground applications that enter ethically problematic terrain is perhaps nowhere as drastic as in the case of the ethics of influence. Generative AI has been connected to the threat of "hypersuasion" (Floridi 2024), or influence that is irresistible, overpowering, and yet cheap and easy to deploy at scale (cf. Klenk 2024).

As social animals, we rely on each other's influence. And we all recognise that others can influence us in good and beneficial ways, but also use their distinctively human capacities, such as empathy, speech, mind-reading, to name just a few, trick us, lead us astray, and make a fool out of us. Generative AI can mimic all of these capacities and thus it invites questions about influence categories once reserved for persons, notably persuasion, manipulation, deception, can now fruitfully be applied to those systems themselves.[8]

Empirical work already suggests that generative models can mislead users, sometimes as an emergent property rather than a designed feature (cf. Hagendorff 2024). The ethical question is not only whether people are mislead by generative AI outputs, but how we should evaluate it when they are, especially when influence arises from interaction without an identifiable human intentention that accounts for it.

A key distinction is between malign and benign influence. Malign forms aim at harming the user in some way, such as by exploitation misinformation. Benign or paternalistic persuasion seeks good ends, such as healthier behaviour, improved learning, emotional well-being, but, crucially, it may still end up using manipulative means (Klenk 2022, 2024). For example, even when the goal is desirable, ethical problems remain if the process bypasses reasoning or consent. A further problem concerns the risk of de-skilling: routine reliance on persuasive systems may lead to cognitive offloading and moral de-skilling, eroding users' capacities for independent judgment.

This debate revives familiar questions from moral psychology and paternalism under new conditions: when, if ever, is it right to influence another's choices for their own good? And can a system that merely simulates understanding ever do so permissibly? Generative AI turns these from abstract philosophical puzzles into urgent design problems.

---

[8] Of course, generative AI can be *used* by humans to deceive, manipulate, or persuade other humans, e.g. by automating well-known scamming tactics or running large-scale influence campaigns. In addition to this, the question is whether these systems' outputs can be analysed in these terms even if there is no explicit intention by a human user associated with the influence.

## 7 Conclusion

Generative AI represents a significant development in how people interact with technological systems. Because these systems can produce outputs that resemble meaningful human expression, they invite forms of engagement that differ from those associated with earlier technologies. I have suggested that this feature can be understood as an affordance: generative AI systems make it natural for users to experience their outputs as if they were intentional or expressive. This affordance does not imply that the systems possess such qualities, but it helps explain the ethical issues that arise when they are used in a wide range of personal and institutional settings.

Attending to this affordance provides a way of addressing several methodological questions about the ethics of generative AI. It identifies a feature that is stable across different applications, that distinguishes generative AI from other forms of machine learning, and that connects the analysis to existing work in moral psychology and the philosophy of technology. It also avoids reliance on speculative claims about future systems, focusing instead on how current systems already influence reasoning, communication, and decision-making.

The ethical implications of generative AI are not uniform. In some contexts, the affordance of experience-as-real may exacerbate familiar concerns about responsibility, privacy, bias, or exploitation. In other contexts, it may create opportunities for new forms of assistance, reflection, or expression. Much depends on how the systems are designed and integrated into social practices. The central point is that ethical evaluation must take account of how users interpret and respond to system behaviour, not only of the systems' internal mechanisms.

Finally, reflection on generative AI may shed light on aspects of human agency and communication that are often taken for granted. When systems can imitate certain forms of expression, questions about the value of authorship, the basis of interpersonal relationships, and the nature of permissible influence arise in new ways. Generative AI thus prompts reconsideration of broader normative questions about how we relate to one another, what we value in our creative and communicative activities, and how these values should guide the design and use of emerging technologies.